\documentclass{article}

\usepackage{graphicx}
\usepackage{amssymb}
\usepackage{lineno}

\usepackage{url}
\usepackage[T1]{fontenc}
\usepackage[utf8]{inputenc}
\usepackage{xspace}
\usepackage{subcaption}
\usepackage{csquotes}
\usepackage{apacite}
\usepackage{natbib}
\usepackage{authblk}
\usepackage{hyperref}
\providecommand{\keywords}[1]{\textbf{\textit{Keywords}} #1}

\usepackage{fancyhdr}
\begin{document}

%\begin{frontmatter}


\title{Natural Language Processing for Music Knowledge Discovery}



%\author{}
\author[1]{Sergio Oramas}
\author[2]{Luis Espinosa-Anke}
\author[3]{Francisco G\'{o}mez}
\author[1]{Xavier Serra}
\affil[1]{Music Technology Group, Universitat Pompeu Fabra}
\affil[2]{School of Computer Science and Informatics, Cardiff University}
\affil[3]{Technical University of Madrid}
\affil[ ]{\textit{\{sergio.oramas,xavier.serra\}@upf.edu, espinosa-ankel@cardiff.ac.uk, fmartin@etsisi.upm.es}}
\maketitle

\begin{abstract}
%Musical knowledge gathered for centuries in written text is leveraged everyday by historians, musicologists and artists who are interested in understanding better the idiosyncrasies and nuances of music. In this scenario, enabling automated access and analysis of textual content is essential for harnessing the potential of written testimonies, especially at large scales. We present different Natural Language Processing (NLP) approaches designed to specifically address challenges of music-related written media, with the ultimate goal to enhance automatic music knowledge discovery. We cover different stages in an NLP pipeline, namely corpus compilation, automatic knowledge graph generation, text-mining, information extraction and sentiment analysis. Each of these approaches is presented alongside different use cases where large collections of documents are processed, and conclusions stemming from a data-driven analysis are showcased and discussed. Specifically, we cover the following use cases: flamenco music (with a corpus of flamenco artist biographies); the Renaissance period (studying a corpus of composers biographies); and a cross-genre diachronic study (analyzing online music reviews).
%In all cases, our approaches deliver a number of meaningful insights, and unveil information which would have otherwise been harder (or impossible) to obtain by traditional scholarly methods.

Today, a massive amount of musical knowledge is stored in written form, with testimonies dated as far back as several centuries ago. In this work, we present different Natural Language Processing (NLP) approaches to harness the potential of these text collections for automatic music knowledge discovery, covering different phases in a prototypical NLP pipeline, namely corpus compilation, text-mining, information extraction, knowledge graph generation and sentiment analysis. Each of these approaches is presented alongside different use cases (i.e., flamenco, Renaissance and popular music) where large collections of documents are processed, and conclusions stemming from data-driven analyses are presented and discussed.
\end{abstract}

%\begin{keyword}
%
%\end{keyword}

%\end{frontmatter}

%\linenumbers


%!TEX root = ../thesis_a4.tex

\section{Introduction}
\label{sec:musicology:introduction}

One of the main tasks carried out in musicology is the development and validation of musical hypotheses. The seed that usually leads to most research involves looking for relevant information in written documents, which in general are organized as compilations, collections or anthologies. Today, it is unsurprising to find many of these collections stored in digitized machine-readable format, a scenario which has signified a great improvement on the way information is accessed. These digitized collections are mostly stored in digital libraries and managed by information systems where documents can be searched by textual keywords. This improvement has increased significantly the possibilities for musicologists to access information. 
However, in these infrastructures the underlying semantics in the textual content of each document are not captured by search engines, which usually operate at an \textit{exact text string matching} level, and therefore in the majority of cases do not take full advantage of the sophisticated processing tools that semantic search puts at our disposal. In this context, and in order to capture the subtle nuances in musical meaning and thus improving musicological research, we argue that it is not enough to put text corpora online and make them searchable. Indeed, there still remains the important and difficult task to transform text collections, from searchable repositories, into knowledge environments, in what can be seen as the next step in the evolution of digital libraries \citep{Fast2011}.

This limitation coexists with an opportunity derived from the quick growth rate at which online content is generated. Today, specifically in the music domain, we have at our disposal vast amounts of knowledge, gathered for centuries by musicologists and music enthusiasts and made accessible by various agents. Most of this knowledge is encoded in artist biographies, reviews, facsimile editions, and other written media. The constant production of this music-related textual information results in large repositories of knowledge, which have great potential for musicological and philological studies. However, since most of it is recorded in \textit{natural language}, processing and analyzing them effectively is a difficult task. We claim, however, that by leveraging \textit{Natural Language Processing} (NLP) techniques, it is possible to unveil relevant information hidden in large domain-specific document collections, which would otherwise remain hidden.

Fortunately, targeting NLP techniques to text corpora in the music domain has been the main focus of several works so far \citep{Oramas2016a,Tata2010,oramas2016sound,
sutcliffe2015,Knees2011,Sordo2012,fujinaga2016digital}. These and other contributions report experimental results of the application of intelligent text processing techniques to music-specific document collections. In addition, many of the upshots of these methods consist in large structured databases containing musical and musicological information, which can provide search engines with much richer and fine-grained information about musicians, their life and work, and even their relation with other musical entities (musicians, record labels, venues, and so forth). Conversely, information already structured in online knowledge repositories has also been exploited in the context of Computational Musicology. For example, a noteworthy example is provided in \cite{Crawford}, where musicologists are provided with a means to create a linked and extensible knowledge structure over a collection of Early Music metadata and facsimile images. In \cite{Rose2014}, seven big datasets of musical and biographical metadata are aligned, showing how analysis and visualization of such data might transform musicological understanding. 
Despite these valuable contributions, scant musicological research has been carried out regarding the specific challenge of processing text collections. 

We propose to specifically address the above challenge by presenting concrete methodologies aimed at exploring large musicological text corpora. With these methodologies we reconcile, on one the hand, intelligent text processing techniques, and on the other, musical knowledge acquired both from structured and unstructured resources. First, we distill methods for gathering and combining information coming from different sources. The textual data used in our experiments comes in different flavors, namely (1) A knowledge base of flamenco music; (2) A corpus of biographies from artists of the Renaissance period; and (3) A dataset of music album reviews of diverse genres.

The underlying knowledge expressed in these corpora is thus extracted applying different NLP pipelines. First, shallow text-mining processing techniques are applied to understand main trends in the different schools of the Renaissance period. Second, Information Extraction (IE) techniques constitute the methodological basis for populating a novel fully automatic flamenco knowledge base, and to analogously study migratory tendencies and the role of different European capitals along the Renaissance period. Third, a methodology for the creation of a knowledge graph from a set of unstructured text documents is proposed and evaluated from different standpoints. We further show a direct application of this knowledge graph in automatically computing the ranked relevance of a given artist in the flamenco and Renaissance corpora. 
Finally, we present an approach for capturing the sentiment expressed in text. Using sentimental information as a starting point, we provide a diachronic study of music criticism via a quantitative analysis of the polarity associated to music album reviews gathered from \texttt{Amazon}\footnote{\url{http://www.amazon.com}}. Our analysis hints at a potential correlation between key cultural and geopolitical events and the language and evolving sentiment found in music reviews and, ultimately, opens exciting avenues for diachronic studies of music genres. 

This paper is an extended version of two previous publications \citep{Oramas2015b,oramas2016exploring}
, with the main novel contributions being the unification of approaches, the addition of more detailed results, and the introduction of an additional use case based on the study of the Renaissance Music period. 
The remainder of this paper is organized as follows. First, in Section~\ref{sec:musicology:corpora}, we describe the processes of gathering and combining information from different data sources, and apply it to the gathering of the three text corpora used throughout the paper: flamenco music, Renaissance artists, and albums reviews. Then, in Section~\ref{sec:musicology:text-mining}, a text-mining approach based on word frequencies is described and applied to study the different music schools of the Renaissance period. Next, in Section~\ref{sec:musicology:information-extraction}, an information extraction pipeline for the extraction of biographical information is exposed and applied to populate a flamenco knowledge base and to study the Renaissance period. Later, in Section~\ref{sec:musicology:knowledge-graphs}, an approach for the creation of knowledge graphs is presented and used to compute a relevance ranking of flamenco and Renaissance artists. In Section~\ref{sec:musicology:sentiment-analysis}, an aspect-based sentiment analysis method is defined and applied to describe how sentiment associated with music reviews changes over time. Following this method, two experiments are performed, one aggregating sentiment scores by review publication year, and other by album publication year. Finally, we conclude with a discussion about our findings (Section~\ref{sec:musicology:conclusions}). 

\section{Collecting text corpora}
\label{sec:musicology:corpora}

Gathering, structuring, and connecting data from different sources is a research problem in itself, where different and major challenges may arise. 
Although some existing repositories with music information, such as Wikipedia\footnote{\url{http://wikipedia.org}}, Oxford Music Online\footnote{\url{http://www.oxfordmusiconline.com/}}, or MusicBrainz\footnote{\url{http://musicbrainz.org}} are quite complete and accurate, there is still a vast amount of music information out there that is generally scattered across different sources on the Web. Selecting the sources and harvesting and combining data is a crucial step towards the creation of practical and meaningful music research corpora \citep{Oramas2014b}

In this work, three different datasets are built as testbeds of our knowledge extraction methodologies. First, we illustrate in detail a methodology for selecting and mixing data coming from different sources in the creation of a flamenco music knowledge base. Then, we apply some of the described approaches to collect a corpus of artist's biographies about Renaissance artists and a collection of music album reviews and metadata. In what follows we describe the gathered corpora and the processes carried on for their compilation.

\subsection{The flamenco corpus}

In this section, we describe the methodology used for the creation of a knowledge base of flamenco music. To this end, a large amount of information is gathered from different data sources, and further combined by applying a process of pair-wise entity resolution. 

\subsubsection{Flamenco music overview}
\label{sec:musicology:flamenco}

Several musical traditions contributed to the genesis of flamenco music as we know it today. Among them, the influences of the Jews, Arabs, and Spanish folk music are recognizable, but indubitably the imprint of Andalusian Gypsies' culture is deeply ingrained in flamenco music. The main components of flamenco music are: \textit{cante} or singing, \textit{toque} or guitar playing, and \textit{baile} or dance. According to \cite{gamboa-05}, flamenco music grew out of the singing tradition, as a melting process of all the traditions mentioned above, and therefore the role of the singer soon became dominant and fundamental. \textit{Toque}  is subordinated to \textit{cante}, especially in more traditional settings, whereas \textit{baile} enjoys more independence from voice. 

In the flamenco jargon, styles are called \textit{palos}. Criteria adopted to define flamenco \textit{palos} are rhythmic patterns, chord progressions, lyrics, poetic structure, and geographical origin. In flamenco, geographical variation is important to classify \textit{cantes} as often they are associated to a particular region where they were originated or where they are performed with gusto. Rhythm or \textit{comp\'as} is a unique feature of flamenco. Rhythmic patterns based on 12-beat cycles are mainly used. Those patterns can be classed as follows: binary patterns, such as \textit{tangos} or \textit{tientos}; ternary patterns, which are the most common ones, such as \textit{fandangos} or \textit{buler\'ias}; mixed patterns, where ternary and binary patterns alternate, such as \textit{guajira}; free-form, where there is no a clear underlying rhythm, such as \textit{ton\'as}. For further information on fundamental aspects of flamenco music, see the book \cite{fer-04}. For a comprehensive study of styles, musical forms and history of flamenco the reader is referred to the books of \cite{bvrr-88}, \cite{nr-95}, and \cite{gamboa-05}, and the references therein.

\subsubsection{Data acquisition}
\label{sec:musicology:datasoruces}

\begin{figure}
	\centering
	\includegraphics[width=0.55\textwidth]{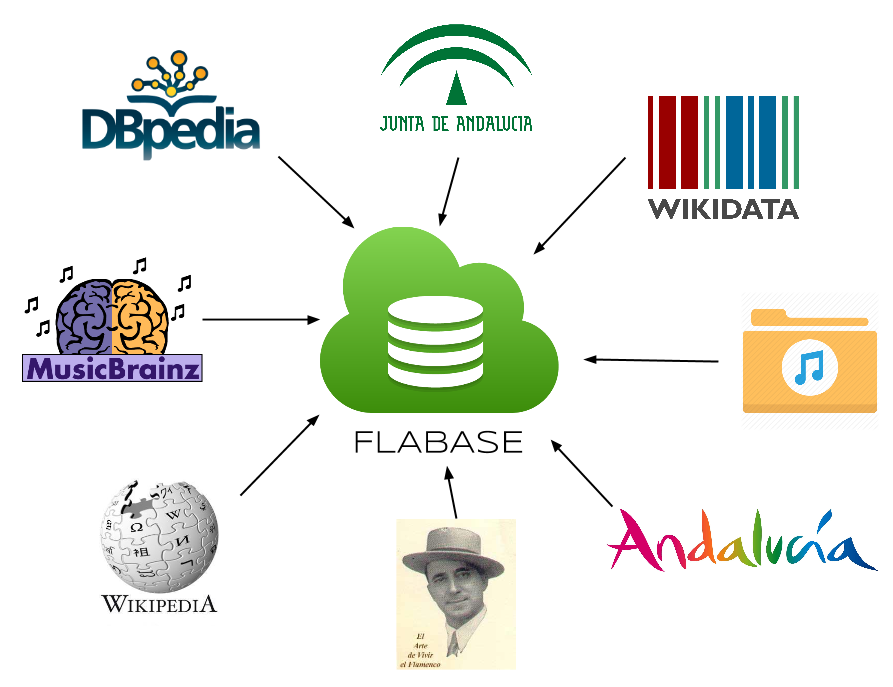}
	\caption{Selected data sources \label{fig:musicology:datasources}}
\end{figure}

Our aim is to gather an important amount of information about musical entities (e.g. artists, recordings), including textual descriptions and available metadata. A schema of the selected data sources is shown in Figure \ref{fig:musicology:datasources}. We started by looking at Wikipedia. 
Each Wikipedia article may have a set of associated categories. Categories are intended to group together pages on similar subjects and are structured in a taxonomical way. To find Wikipedia articles related to flamenco music, we first looked for flamenco categories. The taxonomy of categories can be explored by querying DBpedia, a knowledge base with structured content extracted from Wikipedia. 
We queried the Spanish version of DBpedia\footnote{\url{http://es.dbpedia.org}} for categories related to flamenco. We obtained 17 different categories (e.g., \textit{cantaores de flamenco, guitarristas de flamenco}).

% Figure 1

We gathered all DBpedia resources related to at least one of these categories. We obtained a total number of 438 resources in Spanish, of which 281 were also in English. Each DBpedia resource is associated with a Wikipedia article. Text and HTML code were then extracted from Wikipedia articles in English and Spanish. Next, we classified the extracted articles according to the role of the biography subject (i.e., \textit{cantaor}, guitarist, and \textit{bailaor}). For this purpose, we exploited classification information provided by DBpedia (DBpedia types and Wikipedia categories). At the end, from all gathered resources, we only kept those related to artists and \textit{palos}, totaling  291 artists and 56 \textit{palos}.

As the amount of information present in Wikipedia related to flamenco music is somewhat scarce, we decided to expand our knowledge base with information from two different websites. First, \textit{Andalucia.org}, the touristic web from the Andalusia Government\footnote{\url{http://andalucia.org}}. It contains 422 artist biographies in English and Spanish, and the description of 76 \textit{palos} also in both languages. Second, a website called \textit{El arte de vivir el flamenco}\footnote{\url{http://www.elartedevivirelflamenco.com/}}, which includes 749 artist biographies among \textit{cantaores}, \textit{bailaores} and guitarists. 

We used MusicBrainz to fill our knowledge base with information about flamenco album releases and recordings. For every artist mapped to MusicBrainz, all content related to releases and recordings was gathered. Thus, 814 releases and 9,942 recordings were collected. 

The information gathered from MusicBrainz is a small part of the actual flamenco discography. Therefore, to complement it we used a flamenco recordings database gathered by Rafael Infante and available at CICA website\footnote{\url{http://flun.cica.es/index.php/grabaciones}} (Computing and Scientific Center of Andalusia). This database has information about releases from the early time of recordings until present time, counting 2,099 releases and 4,136 songs. For every song entry, a \textit{cantaor} name is provided, and most of the times also guitarist and \textit{palo}, which is an important piece of information to define flamenco recordings.

Finally, we supplied our knowledge base with information related to Andalusian towns and provinces. We gathered this information from the official database SIMA\footnote{\url{http://www.juntadeandalucia.es/institutodeestadisticaycartografia/sima}} (Multi-territorial System of Information of Andalusia).

\subsubsection{Entity resolution}
\label{sec:musicology:entity_resolution}

Entity resolution is the problem of extracting, matching and resolving entity mentions in structured and unstructured data \citep{Getoor2012}. There are several approaches to tackle the entity resolution problem. For the scope of this research, we selected a pair-wise classification approach based on string similarity between entity labels.

\begin{figure}
	\centering
	\includegraphics[width=0.60\textwidth]{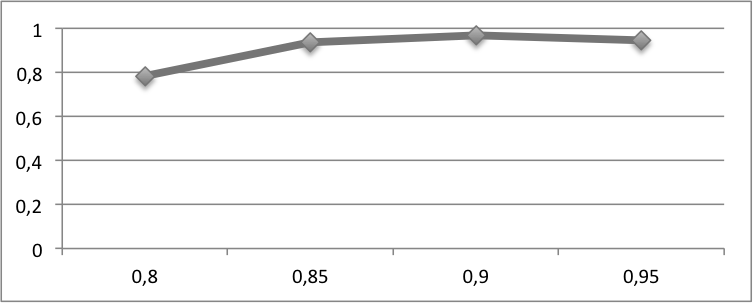}
	\caption{$F$-measure for different values of $\theta$ 
	\label{fig:musicology:fmeasure}}
\end{figure}

The first issue after gathering the data is to decide whether two entities from different sources are referring to the same one. Therefore, given two sets of entities $A$ and $B$, the objective is to define an injective and non-surjective mapping function $f$ between $A$ and $B$ that decides whether an entity $a \in A$ is the same as an entity $b \in B$. To do that, a string similarity metric $sim(a,b)$ based on the Ratcliff-Obershelp algorithm \citep{Ratcliff1988} has been applied. It measures the similarity between two entity labels and outputs a value between 0 and 1. We consider that $a$ and $b$ are the same entity if their similarity is bigger than a parameter $\theta$. If there are two entities $b, c \in B$ that satisfy that $sim(a,b) \geq \theta$ and $sim(a,c) \geq \theta$, we consider only the mapping with the highest score. To determine the value of $\theta$, we tested the method with several $\theta$ values over an annotated dataset of entity pairs. To create this dataset, the 291 artists gathered from Wikipedia were manually mapped to the 422 artists gathered from Andalucia.org, obtaining a total amount of 120 pair matches. As it is shown in Figure~\ref{fig:musicology:fmeasure} the best $F$-measure (0,97) was obtained with $\theta=0.9$. Finally, we applied the described method with $\theta=0.9$ to all gathered entities from the three data sources. Thanks to the entity resolution process, we reduced the initial set of 1,462 artists and 132 \textit{palos} to a set of 1,174 artists and 76 \textit{palos}.

% Figure 2

Once we had our artist entities resolved, we began to gather their related discography. First, we tried to find out the MusicBrainz ID of the gathered artists. Depending on the information about the entity, two different processes were applied. First, we leveraged mapping information between Wikipedia and MusicBrainz present in Wikidata\footnote{\url{https://www.wikidata.org}} \citep{VrandevicandMarkus2014}. Wikidata is a free linked database, which acts as a structured data storage of Wikipedia. For those artists without this mapping information, we queried the MusicBrainz API, and then applied our entity resolution method to the obtained results.

Finally, to integrate the discography database of CICA into our knowledge base, we applied the entity resolution method to the fields \textit{cantaor}, guitarist and \textit{palo} of each recording entry in the database. From the set of 202 \textit{cantaores} and 157 guitarist names present in the recording entries of the database, a total number of 78 \textit{cantaores} and 44 guitarists were mapped to our knowledge base. The number of mapped artists was low due to differences between the way of labeling an artist. An artist name may be written by using one or two of her surnames, or by using her nickname. In the case of \textit{palos}, there were 162 different \textit{palos} in the database, 54 of which were mapped with the 76 of our knowledge base. These 54 \textit{palos} correspond to an 80\% of \textit{palo} assignments present in the recording entries.

\subsubsection{FlaBase}
\label{sec:musicology:flabase}

\begin{figure}
    \centering
    \includegraphics[width=0.60\linewidth]{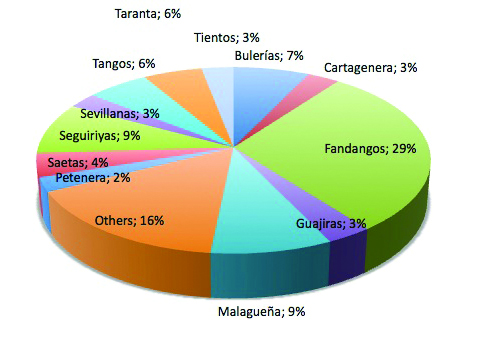}
	\caption{Songs by \textit{palo}}
    \label{fig:musicology:graph-palo}
\end{figure}

FlaBase (Flamenco Knowledge Base) is the acronym of the resulting knowledge base of flamenco music. It contains online editorial, biographical and musicological information related to flamenco music. FlaBase is stored in JSON format, and it is freely available for download\footnote{\url{http://mtg.upf.edu/download/datasets/flabase}}. 
FlaBase contains information about 1,174 artists, 76 \textit{palos} (flamenco genres), 2,913 albums, 14,078 tracks, and 771 Andalusian locations. In Figure~\ref{fig:musicology:graph-palo} it is shown that the most representative \textit{palos} in flamenco music are represented in our knowledge base, with a higher predominance of fandangos.

% Figure 3

\subsection{The Renaissance corpus}
\label{sec:musicology:renaissance-datasets}

Renaissance is a period in history that starts around 1400 with the end of the medieval era, and closes around 1600, with the beginning of the Baroque period. Renaissance music refers to music written in Europe during this period. 

In this work we experimented with two datasets of biographies about Renaissance composers, one gathered from Wikipedia, and another from The New Grove\citep{Germer2001}, available through Oxford Music Online.

\subsubsection{The Wikipedia corpus}
\label{sec:musicology:renaissance-datasets:wikipedia}

In Wikipedia there is an important number of articles related to Renaissance music, most of them biographies of composers. For this research, we compiled the biographies of all composers that are linked in the Wikipedia page: List of Renaissance composers \footnote{\url{https://en.wikipedia.org/wiki/List_of_Renaissance_composers}}. In this page, composers are classified by school. We collected biographies of composers from the five most representative schools: Spanish, German, English, Franco-Flemish, and Italian. A total number of 543 biographies were gathered. In addition to the biography texts, HTML links to other Wikipedia pages present in texts were also stored.

\subsubsection{The Grove corpus}
\label{sec:musicology:renaissance-datasets:grove}

The Grove Dictionary of Music and Musicians \citep{Grove1878} is an encyclopedic dictionary, and one of the largest reference works in Western music. It was first published in four volumes in the last quarter of the XIX century by George Grove. In 1980 a new version called The New Grove \citep{Germer2001} was released with 20 volumes, where there are 22,500 articles and 16,500 biographies. The complete text of the second edition of The New Grove is available in machine-readable format on the online service Oxford Music Online as Grove Music Online. From this set of biographies, we gathered all of them classified as Early Renaissance and Late Renaissance. A total number of 1710 biographies were collected.

\subsection{The albums reviews corpus}

\label{sec:musicology:mard}

In this section, we put forward an integration procedure for enriching with music-related information a large dataset of Amazon customer reviews \cite{McAuley2015a}, with semantic metadata obtained from MusicBrainz. The initial dataset of Amazon customer reviews provides millions of review texts together with additional information such as overall rating (between 0 to 5), date of publication, or creator id. Each review is associated to a product and, for each product, additional metadata is also provided, namely, Amazon product id, list of similar products, price, sell rank, and genre categories. From this initial dataset, we selected the subset of products categorized as \textit{CDs \& Vinyls}, which also fulfill the following criteria. First, considering that the Amazon taxonomy of music genres contains 27 labels in the first hierarchy level, and about 500 in total, we obtain a music-relevant subset and select 16 of the 27 which really define a music style and discard for instance region categories (e.g., World Music) and other categories specifically non-related to a music style (e.g., Soundtrack, Miscellaneous, Special Interest), function-oriented categories (Karaoke, Holiday \& Wedding), or categories whose albums might also be found under other categories (e.g., Opera \& Classical Vocal, Broadway \& Vocalists). We compiled albums belonging only to one of the 16 selected categories, i.e., no multi-label. Note that the original dataset contains not only reviews about CDs and Vinyls, but also about music DVDs and VHSs. Since these are not strictly speaking music audio products, we filter out those products also classified as "Movies \& TV". Finally, since products classified as Classical and Pop are substantially more frequent in the original dataset, we compensate this unbalance by limiting the number of albums of any genre to 10,000. After this preprocessing, the dataset amounts to a total of 65,566 albums and 263,525 customer reviews. A breakdown of the number of albums per genre is provided in Table~\ref{Table1}. The final dataset is called the Multimodal Album Reviews Dataset (MARD) and is freely available for download\footnote{https://www.upf.edu/web/mtg/mard}.

% Table 1
\begin{table}[h]
\centering
\begin{tabular}{l r r r}
\hline
\textbf{Genre} & \textbf{Amazon} & \textbf{MusicBrainz} \\%& \textbf{AcousticBrainz} \\
\hline
Alternative Rock & 2,674 & 1,696 \\%& 564 \\
Reggae & 509 & 260 \\%& 79 \\
Classical & 10,000 & 2,197 \\%& 587 \\
R\&B & 2,114 & 2,950 \\%& 982 \\
Country & 2,771 & 1,032 \\%& 424 \\
Jazz & 6,890 & 2,990 \\%& 863 \\
Metal & 1,785 & 1,294 \\%& 500 \\
Pop & 10,000 & 4,422 \\%& 1701 \\
New Age & 2,656 & 638 \\%& 155 \\
Dance \& Electronic & 5,106 & 899 \\%& 367 \\
Rap \& Hip-Hop & 1,679 & 768 \\%& 207 \\
Latin Music & 7,924 & 3,237 \\%& 425 \\
Rock & 7,315 & 4,100 \\%& 1482 \\
Gospel & 900 & 274 \\%& 33 \\
Blues & 1,158 & 448 \\%& 135 \\
Folk & 2,085 & 848 \\%& 179 \\
\hline
\textbf{Total} & 66,566 & 28,053 \\%& 8,683 \\
\hline
\end{tabular}
\caption{Number of albums by genre with information from the different sources in the albums reviews dataset.}
\label{Table1}
\end{table}

Having performed genre filtering, we enrich the dataset by extracting artist names and record labels from the Amazon product page. We pivot over this information to query the MusicBrainz search API to gather additional metadata such as release id, first release date, song titles and song ids. Mapping with MusicBrainz is performed using the same methodology described in Section~\ref{sec:musicology:entity_resolution}, following a pair-wise entity resolution approach based on string similarity with a threshold value of $\theta=0.85$. We successfully mapped 28,053 albums to MusicBrainz. 

\section{Text-mining}
\label{sec:musicology:text-mining}

Text-mining is the process of deriving high-quality information from text. This high-quality information is typically derived through the devising of patterns and trends using statistical analysis over text. Many text-mining techniques are based on the analysis of frequencies of the words present in the set of studied documents. In what follows, we illustrate the potential of this technique with a simple application to the analysis of word frequencies in our corpus of Renaissance artist's biographies. 

\subsection{Renaissance music schools}

\begin{figure}[ht!]
    \centering
    \begin{subfigure}{.49\textwidth}
        \centering
        \includegraphics[width=\linewidth]{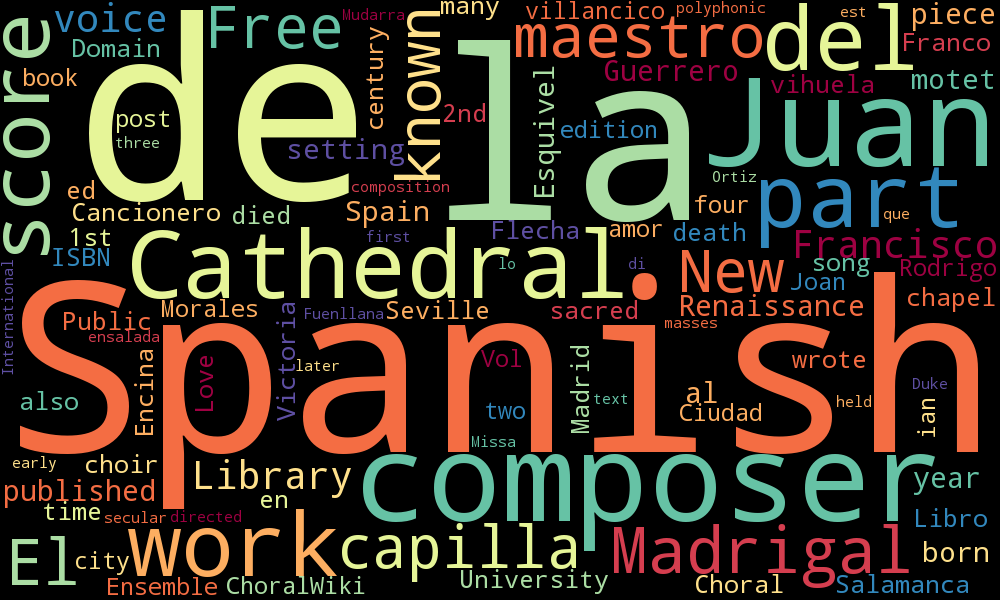}
        \caption{Spanish}
        \label{fig:musicology:spanish}
		\vspace*{2mm}
    \end{subfigure}
    \begin{subfigure}{.49\textwidth}
        \centering
        \includegraphics[width=\linewidth]{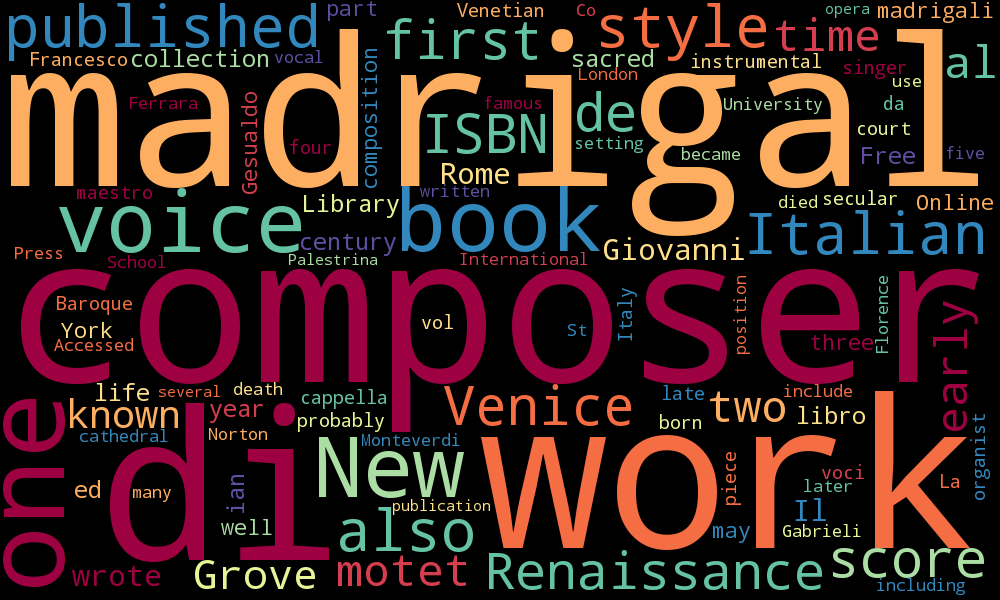}
        \caption{Italian}
        \label{fig:musicology:italian}
		\vspace*{2mm}
	\end{subfigure}
    \begin{subfigure}{.49\textwidth}
        \centering
        \includegraphics[width=\linewidth]{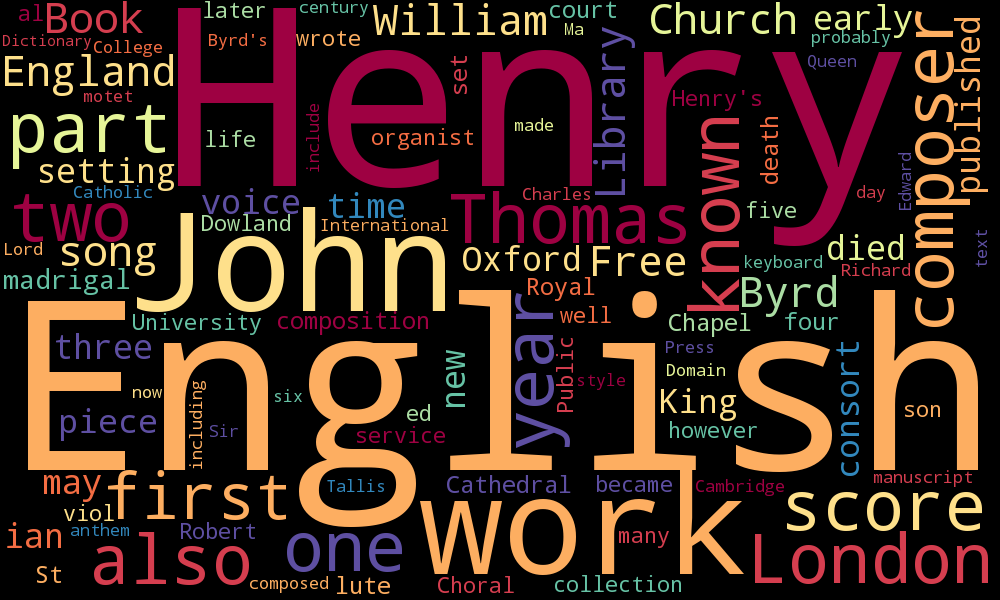}
        \caption{English}
        \label{fig:musicology:english}
		\vspace*{2mm}
    \end{subfigure}
    \begin{subfigure}{.49\textwidth}
        \centering
        \includegraphics[width=\linewidth]{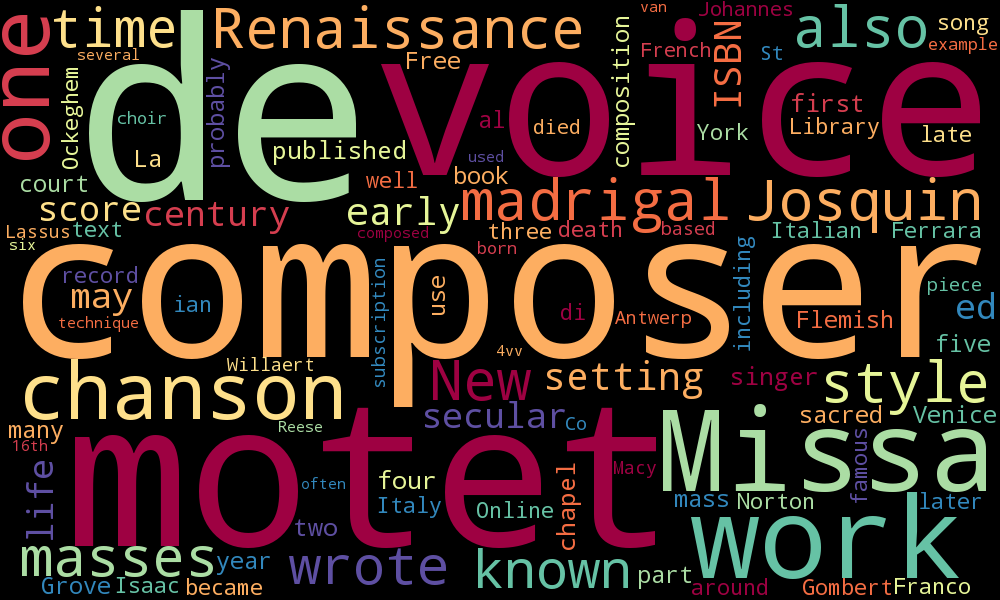}
        \caption{Franco-Flemish}
        \label{fig:musicology:franco-flemish}
		\vspace*{2mm}
	\end{subfigure}
    \begin{subfigure}{.49\textwidth}
        \centering
        \includegraphics[width=\columnwidth]{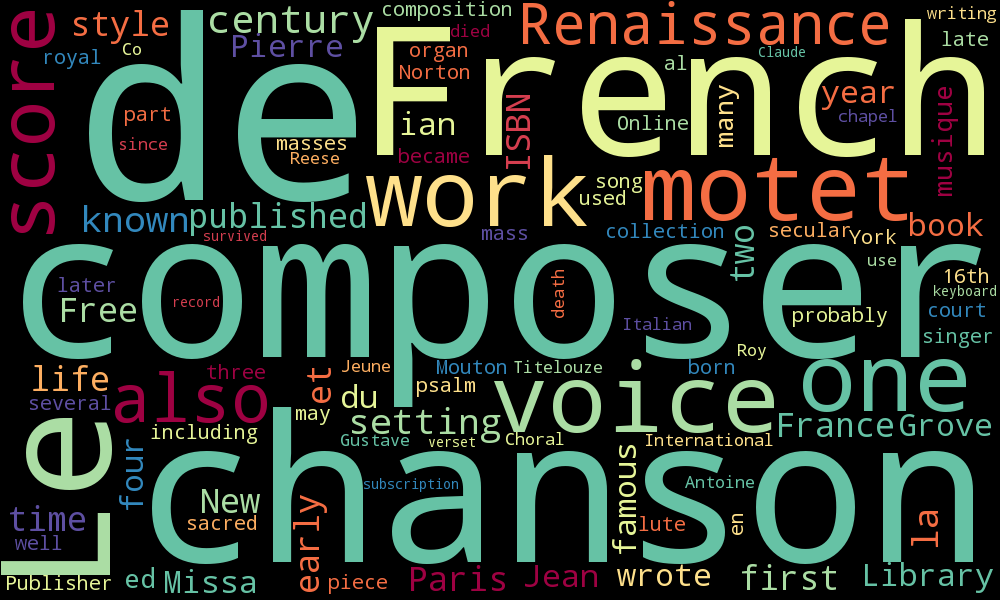}
        \caption{French}
        \label{fig:musicology:french}
    \end{subfigure}
    \begin{subfigure}{.49\textwidth}
        \centering
        \includegraphics[width=\linewidth]{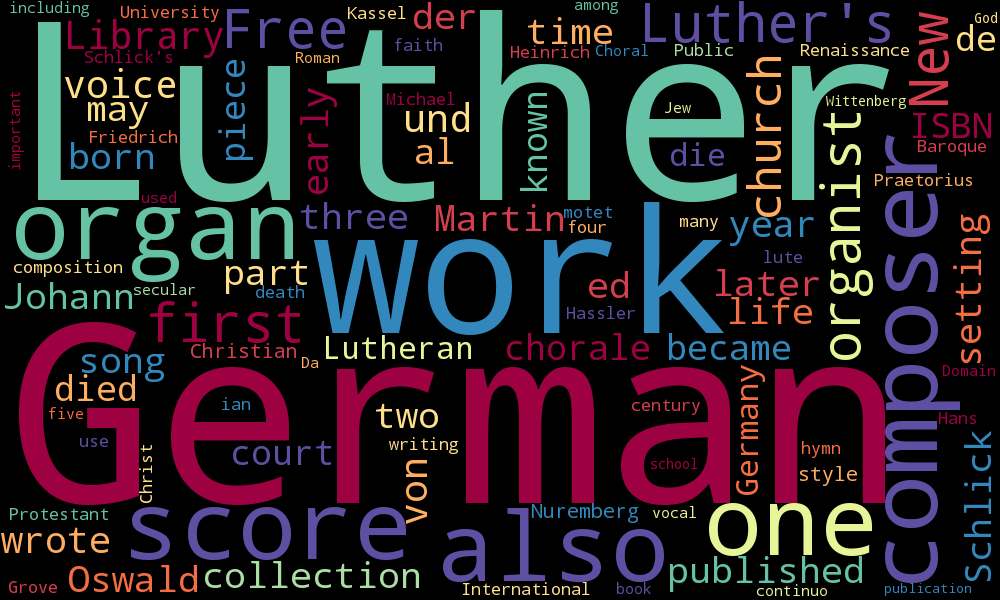}
        \caption{German}
        \label{fig:musicology:german}
    \end{subfigure}
    \caption{Word clouds by school from Wikipedia biographies.}
    \label{fig:musicology:wordclouds}
\end{figure}

The computational analysis of artist biographies may reveal interesting insights from the data that can be useful to musicologists. Using the Renaissance artist's biographies gathered from Wikipedia (see Section~\ref{sec:musicology:renaissance-datasets:wikipedia}), we applied a shallow analysis of the words used in the articles. We computed the frequencies of all words present in the articles of every school. From the obtained frequencies we plot a word cloud for each school, where more frequent words are represented with bigger fonts. In Figure~\ref{fig:musicology:wordclouds}, the word clouds of the different schools are shown. We observe very clear insights from the images at first sight. We see, for instance, how madrigal is very important in the Italian, chanson in the French, and motet in the Franco-Flemish school. We also see the importance of the Church in the Spanish school, or the relevance of organ music in the German school. Although these observations may seem obvious to a musicologist, they are extracted directly from the data without human intervention. This approach can be applied to text corpora the researcher might not be familiar with, helping her in easily discovering some trends directly from the data.

% Figure 4

\section{Information Extraction}
\label{sec:musicology:information-extraction}

Information extraction is the task of automatically extracting structured information from unstructured or semi-structured text sources. It is a widely studied topic within the NLP research community \citep{cowie1996information}.
A major step towards understanding language is the extraction of meaningful terms (entities) from text as well as relationships between those entities. This statement involves two different tasks. First, the identification and categorization of entity mentions. This task is called named entity recognition (NER). However, when this task involves a latter step of disambiguation of entities against a knowledge base it is called named entity disambiguation or entity linking (EL). The second task consists of the identification of relevant semantic relations or attributes associated to these entities. 

\subsection{Entity linking}\label{sec:musicology:entity_linking}

The advent of large knowledge repositories and collaborative resources has contributed to the emergence of entity linking, i.e., the task of discovering mentions of entities in text and link them to a suitable knowledge repository \citep{Moroetal2014}. 
It encompasses similar subtasks such as named entity disambiguation \citep{BunescuandPasca2006}, which is precisely linking mentions of entities to a knowledge base, or wikification \citep{MihalceaandCsomai2007}, specifically using Wikipedia as knowledge base.

Entity linking is typically divided in two steps, namely, the identification of a text span from the text as an entity candidate, and the disambiguation of this entity with respect to a knowledge base. This disambiguation step can be directly applied to the surface form of the identified text span, or to the output of a NER system previously applied. The biggest difference here is that the NER system not only identifies the text span, but also provides a category that classify the identified candidate. We propose a method that employs a combination of both approaches, depending on the category of the entity. For NER, we used the Stanford NER system \citep{Finkel2005}, implemented in the library Stanford Core NLP\footnote{\url{http://nlp.stanford.edu/software/corenlp.shtml}} and trained on English and Spanish texts. For disambiguation we simply looked for exact string matches between entity labels in the knowledge base and identified text spans. 

\subsection{Studying the flamenco corpus}

\subsubsection{Linking entities in the flamenco corpus}

As the gathered flamenco texts are mostly written in Spanish, we needed an entity linking system that deals with Spanish texts. Although there are many entity linking tools available, state-of-the-art systems are well-tuned for English texts, but may not perform as well in languages other than English, and even less with music related texts~\citep{Oramas2016}. In addition, we wanted to have a system that uses our own knowledge base for disambiguation. Therefore, we developed our own system, which is able to detect and disambiguate three categories of entities: Person, \textit{Palo} and Location. Three different approaches for the selection of annotation candidates were defined by applying NER only on a subset of the categories of entities: only using text spans (no NER) for disambiguation; disambiguating Location and Person entities from the NER output, and \textit{Palo} from text spans; and only disambiguating Location entities from the NER output, and Person and \textit{Palo} directly from text spans.

To determine which approach performs better, three artist biographies were manually annotated, having a total number of 49 annotated entities. Results on the different approaches are shown in Table~\ref{Table2}. We observe that applying NER to entities of the Person category worsens performance significantly, as recall suddenly decreases by half. After manually analyzing false negatives, we observed that this is caused because many artist names have definite articles between name and surname (e.g., \textit{de, del}), and this is not recognized correctly by the NER system. In addition, many artists have a nickname that is not interpreted as a Person entity by the NER system. The best approach is the third one (NER to LOC), where NER output is used only for Locations, which is slightly better than the first one (no NER) in terms of precision. This is due to the fact that many artists have a town name as a surname or as part of his or her nickname. Therefore, applying entity linking directly to text spans is misclassifying Person entities as Location entities. Thus, by adding a previous step of NER to Location entities we have increased the overall performance, as it can be seen on the $F$-measure values.

% Table 2
\begin{table}
	\centering
	  \begin{tabular}{  l c c c }
    \hline
    Approach & Precision & Recall & $F$-measure \\ 
    \hline
    1) no NER & 0.829 & \textbf{0.694} & 0.756 \\ 
    2) NER to PERS \& LOC & 0.739 & 0.347 & 0.472 \\
    3) NER to LOC & \textbf{0.892} & 0.674 & \textbf{0.767} \\
    \hline
  \end{tabular}
	\caption{Precision, Recall and $F$-measure of entity linking approaches.}
	
	\label{Table2}
\end{table}

\subsubsection{Extracting biographical data}
\label{sec:musicology:ie}

While the created knowledge base of flamenco does already encode relevant culture and music-specific information, a notable portion of the data collected currently remains unexploited due to its unstructured nature. Consequently, to enhance the amount of structured data, a process of information extraction is carried out. We focus on extracting two specific pieces of information from the artist biographies: birth year and birth place, as they can be relevant for anthropological studies. We observed that this information is often in the first sentences of the biographies, and always near the word \textit{naci\'{o}} (Spanish translation of "was born"). Therefore, to extract this information, we look for this word in the first 250 characters of every biographical text. If it is found, we apply our entity linking method to this piece of text. If a Location entity is found near the word "naci\'{o}", we assume that this entity is the place of birth of the biography subject. In addition, by using regular expressions, we look for the presence of a year expression in the context of the Location entity. If it is found, we assume it as the year of birth. If more than one year is found, we select the one with the smaller value. 

To evaluate our approach, we tested the extraction of birth places in all texts coming from the web Andalucia.org (442 artists). We manually annotated the province of provenance of these 442 artists for building ground truth data. After the application of the extraction process on the annotated test set, we obtained a precision value of 0.922 and a recall of 0.648. Therefore, we may argue that our method is extracting biographic information with high precision and quite reasonable recall. We finally applied the extraction process to all artist entities with biographical texts. Thus, 743 birth places and 879 birth years were extracted. 

\begin{figure}[]
    \centering
    \begin{subfigure}{.45\textwidth}
        \centering
        \includegraphics[width=.9\linewidth]{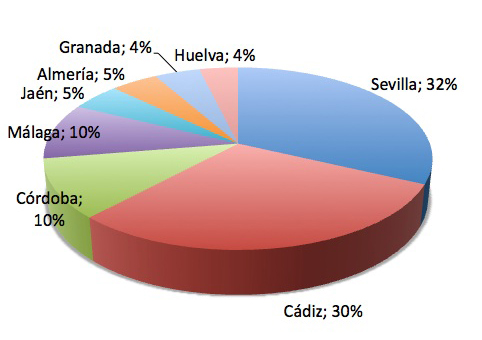}
		\caption{Artists by province of birth}
		\label{fig:musicology:graph-province}
    \end{subfigure}
    \begin{subfigure}{.45\textwidth}
        \centering
        \includegraphics[width=6cm]{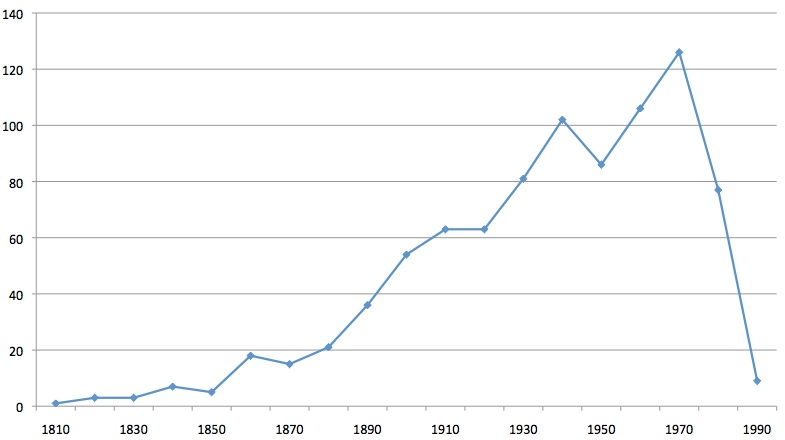}
        \caption{Artists by decade of birth} 
        \label{fig:musicology:graph-decade}
    \end{subfigure}
    \caption{FlaBase distributions.}
    \label{fig:musicology:graphs-flabase}
\end{figure}

Using the information extracted, we computed the distribution of different items present in FlaBase. Data shown in Figures~\ref{fig:musicology:graph-province} and \ref{fig:musicology:graph-decade} was obtained thanks to the information extraction process applied. We can observe in Figure~\ref{fig:musicology:graph-province} that most flamenco artists are from the Andalusian provinces of Seville and Cadiz. 
Finally, in Figure~\ref{fig:musicology:graph-decade} we observe a higher number of artists in the data were born from the 30's to the 80's of the 20th century.

% Figure 5

\subsection{Studying the Renaissance period}

\begin{figure}[ht!]
    \centering    
    \begin{subfigure}{.49\textwidth}
        \centering
        \includegraphics[width=\linewidth]{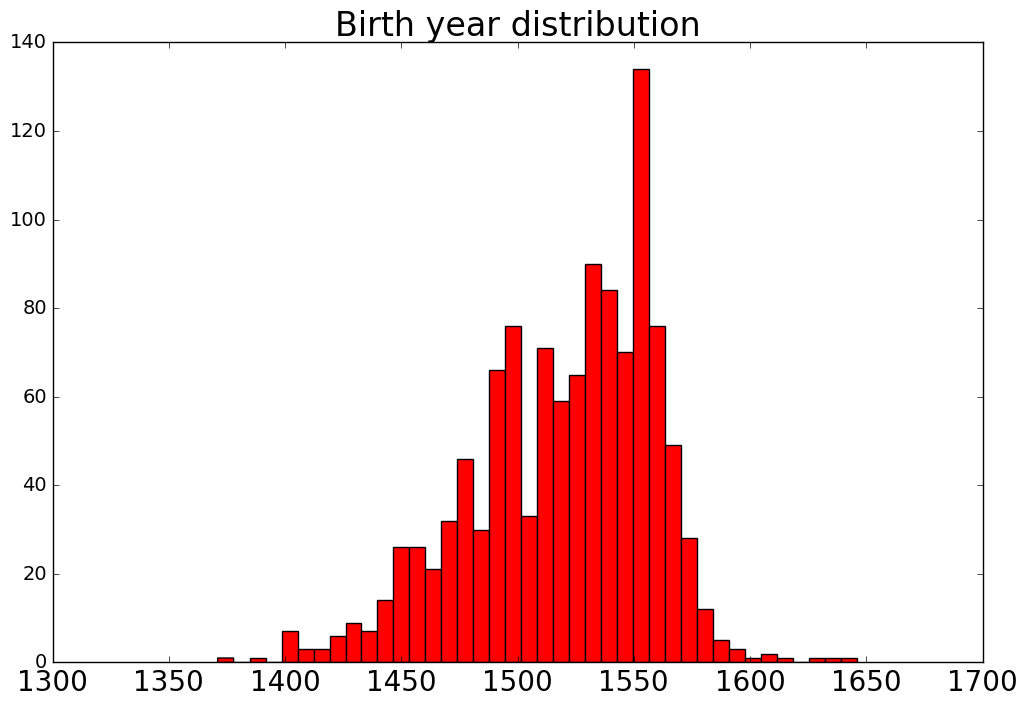}
        \caption{Births}
        \label{fig:musicology:births}
    \end{subfigure}
    \begin{subfigure}{.49\textwidth}
        \centering
        \includegraphics[width=\linewidth]{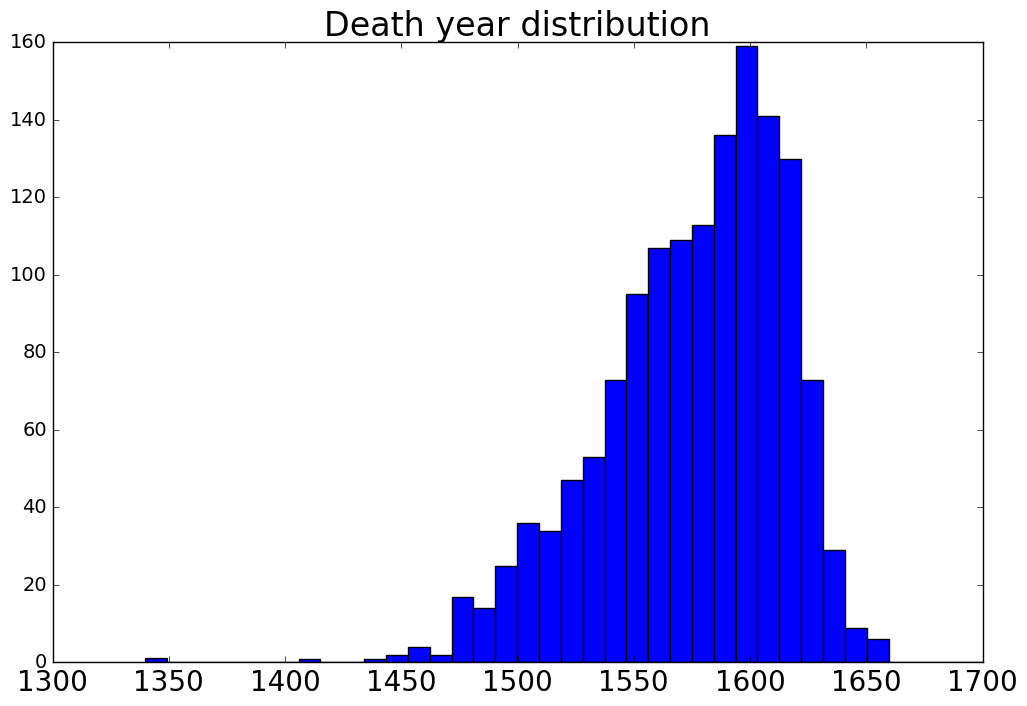}
        \caption{Deaths}
        \label{fig:musicology:deaths}
    \end{subfigure}
    \caption{Distribution of birth and death dates.}
    \label{fig:musicology:birth-death}
\end{figure}

To study the Renaissance period, we applied a process of information extraction similar to the one described above for the flamenco corpus. Thus, we extracted biographical data from the artist biographies in the Grove corpus. We observed in this corpus that at the beginning of every biography there is a sentence between parentheses with information about the place and date of birth and death. Therefore, we automatically extracted this information using the same ad-hoc entity linking system and regular expressions used to extract information from the flamenco corpus. Using the extracted data, we first plotted the histograms of the distributions of birth and death dates (Figures~\ref{fig:musicology:births} and \ref{fig:musicology:deaths}. As observed in Figures~\ref{fig:musicology:births} and \ref{fig:musicology:deaths}, most Renaissance composers were born in the first half of the XVI century, and died at the beginning of the XVII century. This image gives as a simple overview of the activity in the period.

% Figure 6

Using the extracted places of birth and death, we also computed the difference between cities in number of births and deaths. We observe in Table~\ref{Table3} that Brescia and Parma are cities where many relevant composers were born, but few died. This perhaps implies a good educational environment in music, but less career opportunities for those composers. By contrast, we observe in Table~\ref{Table4} how big cities like Rome, London, Paris, or Venice are attractors of talent, with much larger number of deaths than births. Florence in contrast, typically considered as the cradle of the Renaissance, has a similar number of births and deaths.

% Table 3
\begin{table}[]
\centering
\begin{tabular}{l r r r}
\hline
\textbf{City} & \textbf{Births} & \textbf{Deaths} & \textbf{Difference} \\
\hline
Florence & 25 & 26 & +4\% \\
Brescia & 18 & 5 & -72\% \\
Parma & 15 & 10 & -33\% \\
Nuremberg & 15 & 17 & +13\% \\
Bologna & 15 & 13 & -13\% \\
\hline
\end{tabular}
\caption{Top cities by number of births, extracted from the Grove dataset.}
\label{Table3}
\end{table}

% Table 4
\begin{table}[]
\centering
\begin{tabular}{l r r r}
\hline
\textbf{City} & \textbf{Births} & \textbf{Deaths} & \textbf{Difference} \\
\hline
Rome & 9 & 59 & +555\% \\
London & 7 & 39 & +457\% \\
Paris & 6 & 32 & +433\% \\
Venice & 11 & 29 & +164\% \\
Florence & 25 & 26 & +4\% \\
\hline
\end{tabular}
\caption{Top cities by number of deaths, extracted from the Grove dataset.}
\label{Table4}
\end{table}

Finally, we computed the median of the distribution of death years by city of those with larger number of deaths. This data may be useful to observe when a city was in the middle of his success as an attractor of musical talent. In Table~\ref{Table5}, we observe how the gravity center of Renaissance music moves from Nuremberg and Paris to Venice, Florence, and Rome, and finally to London. Again, this result may be very illustrative as a first impression of this musical period.

% Table 5
\begin{table}[]
\centering
\begin{tabular}{l r}
\hline
\textbf{City} & \textbf{Median year} \\
\hline
Nuremberg & 1563 \\
Paris & 1569 \\
Venice & 1576 \\
Rome & 1594 \\
Florence & 1597 \\
London & 1610 \\
\hline
\end{tabular}
\caption{Median of the distribution of deaths by city.}
\label{Table5}
\end{table}

\section{Knowledge graph construction}
\label{sec:musicology:knowledge-graphs}

We assume that an entity mention inside an artist biography signals a semantic relation between the entity that constitutes the main theme of the biography (subject entity) and the mentioned entity. Based on this assumption, we build a semantic graph by applying the following steps. First, each artist in the corpus is added to the graph as a node. Second, entity linking is applied to artist's biographical texts. For every linked entity identified in the biography, a new node is created in the graph (only if it was not previously created). Next, an edge is added, connecting the subject entity with the linked entity found in its biography. This way, a directed graph connecting the entities of the text corpus is  obtained. 

This graph may have multiple applications. It may be exploited to compute similarity measures between artists, as explored in~\cite{Oramas2015a}, or it may provide a data structure well suited to the implementation of graphical navigational systems throughout the collection of documents, as explored in \cite{Oramas2014}. In this work, we explore a different application: the measurement artist relevance. 

\subsection{Artists relevance}
\label{sec:musicology:relevance}

Entities identified in a text by an entity linking system may be seen as hyperlinks that connects one text to another. Thus, algorithms to measure the relevance of nodes in a network of hyperlinks can be applied to our semantic graph \citep{Bellomi2005}. Hence, a knowledge graph constructed with the proposed methodology represents a network of hyperlinks that connect the different documents in the corpus. In order to measure artist relevance in our constructed graph, we applied the PageRank \citep{Brin1998} and HITS \citep{Kleinberg1999} algorithms. PageRank outputs a measure of relevance for each node, and HITS gives two different results: \textit{authority} and \textit{hubness}. We only take into consideration \textit{authority} from HITS algorithm because it has been proven to be the most effective of both values as a metric of relevancy~\citep{Bellomi2005}. 

\subsubsection{Flamenco artists}
\label{sec:musicology:flamenco-relevance}

Following the proposed methodology for the creation of a knowledge graph, we created a graph of flamenco artists after its application to the corpus of artist biographies gathered in FlaBase. We applied the entity linking system described in Section~\ref{sec:musicology:entity_linking} and then constructed the graph. In this case, we also added other attributes present in FlaBase to the graph, such as the extracted attributes and the recordings associated to each artist. Once the graph was built, we applied the PageRank and HITS algorithms and built an ordered list with the top-10 entities of the different artist categories (\textit{cantaor}, guitarist and \textit{bailaor}) for each of the algorithms. 

For evaluation purposes, we asked a reputed flamenco expert to build a list of top-10 artists for each category according to his knowledge and the available bibliography. The concept of artist relevance is somehow subjective and there is no unified or consensual criterion for flamenco experts about who the most relevant artists of all time are. Despite that, there is a high level of agreement among them on certain artists that should be on such a hypothetical list, based on their influence in the evolution of the genre. Thus, after consulting several documented sources and other flamenco experts, our expert provided us with this list of consensual top-10 artists by category and we considered it as ground truth. 

We define precision as the number of identified artists in the resulting list that are also present in the ground truth list divided by the length of the list. We evaluated the output of the two algorithms by calculating precision over the entire list (top-10), and over the first five elements (top-5) (see Table~\ref{Table6}). We can observe that PageRank results show the greatest agreement with the flamenco experts list. 
High values of precision, especially for the top-5 list, indicates that the information in the knowledge graph is highly complete and accurate, and the proposed methodology adequate to compute relevance of artists. In Table~\ref{Table7} the top-5 artists in each category obtained with the PageRank algorithm are shown. 
It is clear that this approach tend to favor ancient artists that have more probabilities of have been mentioned in other biographies. Therefore, we understand in this case artist relevance as a measure directly tied to the influence of the artist in the evolution of the genre.

% Table 6
\begin{table}[]
    \centering
    \begin{tabular}{ c c c }
    \hline
    & Top-5 & Top-10 \\
    \hline
    PageRank & 0.933 & 0.633 \\
    HITS Authority & 0.6 & 0.4 \\
    \hline
    \end{tabular}
    \caption{Precision values of artist relevance ranking.}    
    \label{Table6}
\end{table}

% Table 7
\begin{table}[]
    \centering
    \begin{tabular}{c c c }
    \hline
    \textit{Cantaor} & Guitarist & \textit{Bailaor} \\
    \hline
Antonio Mairena & Paco de Luc\'{i}a & Antonio Ruiz Soler \\
Manolo Caracol & Ram\'{o}n Montoya & Rosario \\
La Ni\~{n}a de los Peines & Ni\~{n}o Ricardo & Antonio Gades \\
Antonio Chac\'{o}n & Manolo Sanl\'{u}car & Mario Maya \\
Camar\'{o}n de la Isla & Sabicas & Carmen Amaya \\
Manuel Torre & Tomatito & Pilar L\'{o}pez \\
José Merc\'{e} & Vicente Amigo & La Argentinita \\
Enrique Morente & Gerardo N\'{u}\~{n}ez & Lola Flores \\
Pepe Marchena & Paco Cepero & Pastora Imperio \\
Manuel Vallejo & Pepe Habichuela & Jos\'{e} Antonio \\
    \hline
    \end{tabular}
    \caption{PageRank Top-10 artists by category.}    
    \label{Table7}
\end{table}

\subsubsection{Renaissance artists}
\label{sec:musicology:renaissance-relevance}

\begin{figure}[!ht]
	\centering
	\includegraphics[width=8cm]{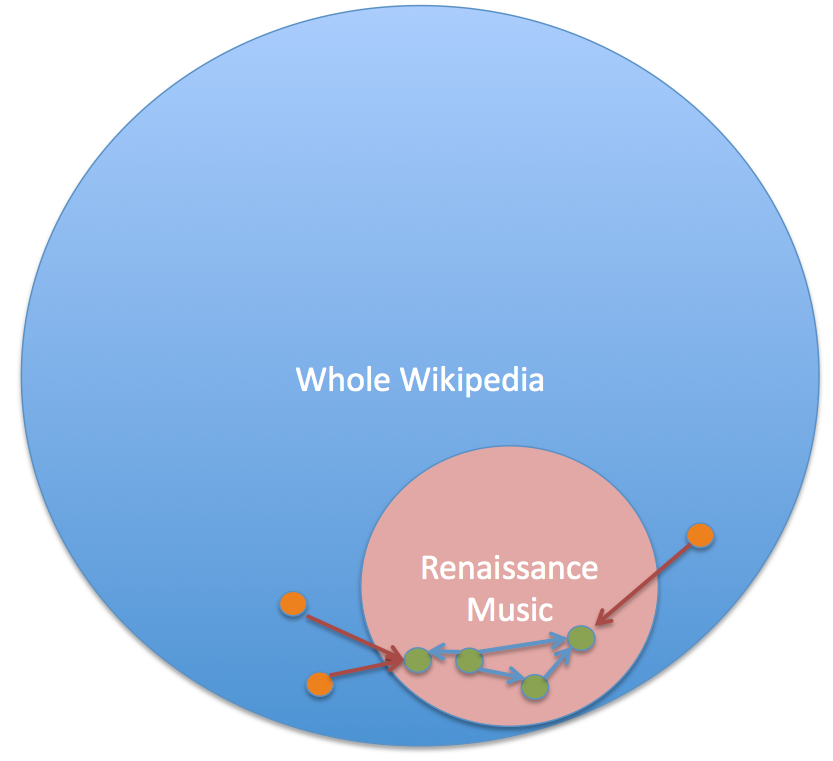}
	\caption{Knowledge graph construction approaches 
	\label{fig:musicology:wiki-graph}}
\end{figure}

We followed two different strategies for knowledge graph construction for the two datasets of Renaissance artist biographies. For the Wikipedia corpus, we took advantage of the links already present in the Wikipedia pages instead of applying entity linking. We connected the biography main theme entity with all the entities linked in the biography text. There are many entities linked in the biographies that do not correspond to Renaissance composers (e.g., countries, events, kings). Therefore, we created a graph composed only of Renaissance composers and another with all the entities found in the biographies. Figure~\ref{fig:musicology:wiki-graph} shows the difference between the two graphs.

% Figure 7

% Table 8
\begin{table}[]
\centering
\begin{tabular}{l l l}
\hline
\textbf{School} & \textbf{Internal connections} & \textbf{All connections} \\
\hline
Spanish & Francisco Guerrero & Juan de la Encina \\
German & Hans Leo Hessler & Martin Luther \\
English & Thomas Morley & Henry VIII \\
Franco-Flemish & Josquin des Prez & Josquin des Prez \\
Italian & Palestrina & Monteverdi \\
\hline
\end{tabular}
\caption{Relevance ranking of composers by school and graph creation approach using the Wikipedia dataset.}
\label{Table8}
\end{table}

% Table 9
\begin{table}[]
\centering
\begin{tabular}{l l l}
\hline
\textbf{Ranking} & \textbf{Internal connections} & \textbf{All connections} \\
\hline
\#1 & Josquin des Prez & Henry VIII \\
\#2 & Palestrina & Martin Luther \\
\#3 & Orlande de Lassus & Henry V \\
\#4 & Adrian Willaert & Monteverdi \\
\hline
\end{tabular}
\caption{Relevance ranking of all composers by graph creation approach using the Wikipedia dataset.}
\label{Table9}
\end{table}

Following the same methodology described in Section~\ref{sec:musicology:relevance}, we computed the relevance ranking of the composers in the 2 graphs created from the Wikipedia corpus using the PageRank algorithm. 
We observe in Table~\ref{Table8} the most relevant composer of each school obtained from the 2 Wikipedia graphs, the one using only links between Renaissance composers (\textsc{internal connections}) and the one using links to any entity (\textsc{all connections}). From a musicological perspective, we observe that the results using only \textsc{internal connections} have more sense than those obtained using \textsc{all connections}. For example, Henry VIII appears as the most prominent entity of the English school when using \textsc{all connections}. Henry VIII, in addition to the king of England, was a composer of the Renaissance era. However, his popularity is mainly due to his role as a king rather than as a composer. Using \textsc{internal connections} only we obtain Thomas Morley as the most prominent composer of the English school, who is really a cornerstone of this school. The same happens in the German school with Martin Luther, who is popular for other aspects different from music. In the Italian school we observe a slightly different situation. Claudio Monteverdi appears as the most prominent composer using \textsc{all connections}. He is actually one of the most prominent composers of the history of music, however, although he started his career in the Renaissance era, he is mostly considered as a Baroque composer. Palestrina, who was obtained using only \textsc{internal connections}, is also a very prominent composer in the history of music, but he is a prototypical composer of the Renaissance. We can infer from these results that the use of only inner connections helps the approach to obtain results that are more musicologically meaningful. In Table~\ref{Table9} we observe the top-4 composers from both graphs independently of the music school. We notice here the same tendency in the results.

% Table 10
\begin{table}[]
\centering
\begin{tabular}{l l}
\hline
\textbf{Ranking} & \textbf{Internal connections} \\
\hline
\#1 & Palestrina \\
\#2 & Alessandro Damasceni Peretti di Montalto \\
\#3 & Petrarch\\
\#4 & Claudin de Sermisy \\
\#5 & Luca Marenzio\\
\#6 & Pierre Sandrin\\
\#7 & Jaques Arcadelt\\
\#8 & Jacob Obrecht\\
\hline
\end{tabular}
\caption{Relevance ranking of all composers using the Grove dataset.}
\label{Table10}
\end{table}

For the Grove corpus, we followed the same strategy described for the construction of the flamenco knowledge graph. We applied entity linking to the biographies, and then connected each biography subject with the entities mentioned in its biography. We employed a similar ad-hoc approach for entity linking as the one described in Section~\ref{sec:musicology:entity_linking}. 

We computed the ranking list of the most relevant composers applying the PageRank algorithm over this graph. As shown in Table~\ref{Table10}, composers obtained in this list are all very relevant musicians of the Renaissance period. However, there are other types of artists and relevant people in the list, similarly to the results obtained with the Wikipedia graph with \textsc{all connections}. This implies the need of a filtering process of the selected entities. This result confirms the findings shown in Section~\ref{sec:musicology:flamenco-relevance}, which demonstrate the utility of the proposed approach to compute artist relevance ranking from unstructured texts by using entity linking.

\section{Sentiment analysis}
\label{sec:musicology:sentiment-analysis}

Sentiment analysis is the task to systematically identify, extract, quantify, and study affective states and subjective information in text. 
Among the different subtasks of sentiment analysis, we focus in this work on aspect-based sentiment analysis. This technique provides specific sentiment scores for different aspects present in the text, e.g. album cover, guitar, voice or lyrics. These scores represent how much the user likes or dislikes specific attributes expressed in text.

\subsection{Aspect-based sentiment analysis}

% Figure 8
\begin{figure}
\includegraphics[width=\columnwidth]{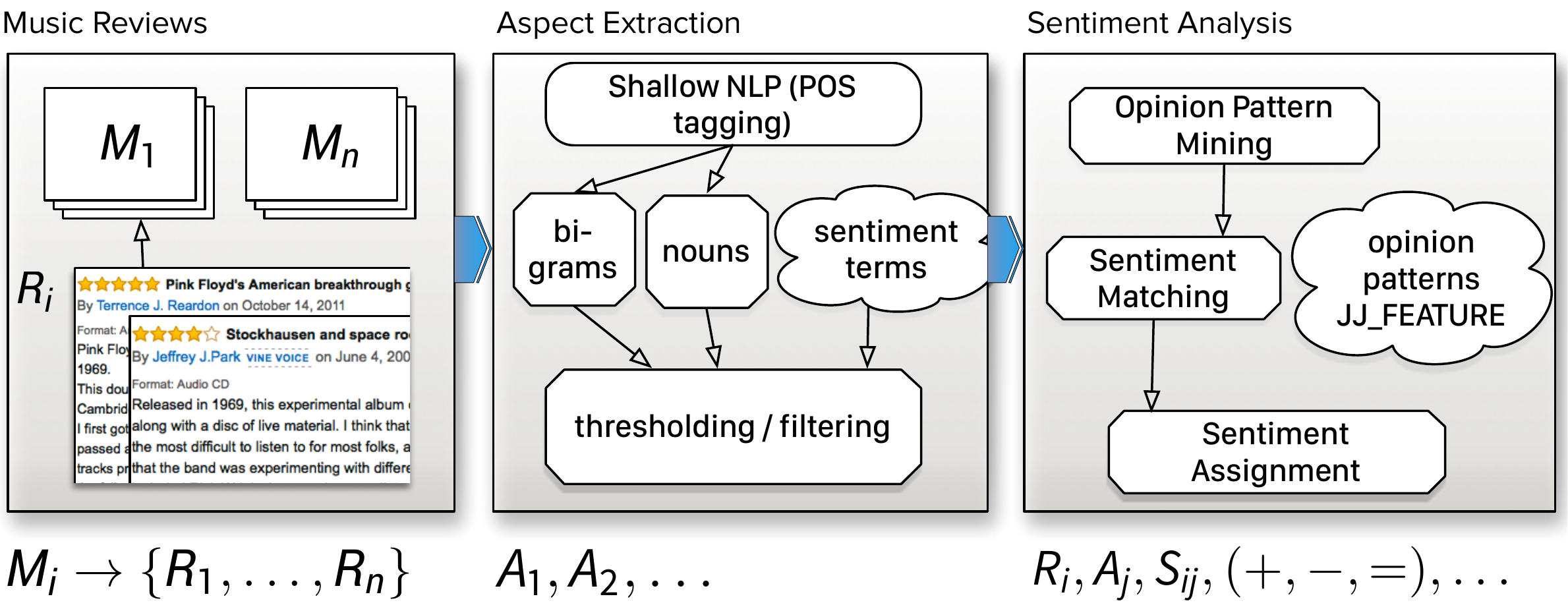}
\caption{Overview of the opinion mining and sentiment analysis framework.}
\label{fig:musicology:OMF}
\end{figure}

Following the work of \cite{DongSOS13,DongOS14} we use a combination of shallow NLP, opinion mining, and sentiment analysis to extract opinionated features from reviews. For all reviews $R_{i}$ of each album, we mine bi-grams and single-noun aspects (also called review features; see \cite{Hu2004}). We consider bi-grams that conform to a noun followed by a noun (e.g., \emph{chorus arrangement}) or an adjective followed by a noun (e.g., \emph{original sound}), and excluded bi-grams whose adjective is a sentiment word (e.g., \emph{excellent}, \emph{terrible}). Separately, single-noun aspects are validated by eliminating nouns that are rarely associated with sentiment words in reviews, since such nouns are unlikely to refer to item aspects. We refer to each of these extracted aspects $A_{j}$ as review aspects.

\begin{figure}[h]
\includegraphics[width=\columnwidth]{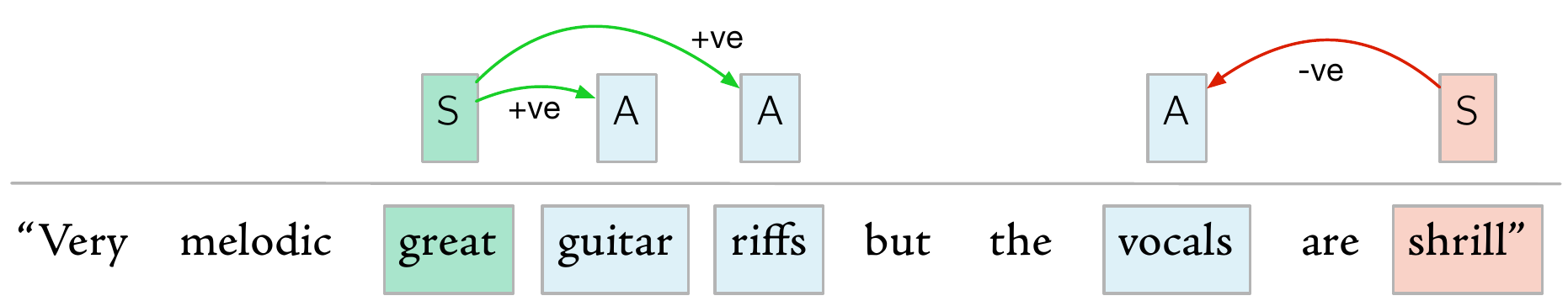}
\caption{A sentence from a sample review annotated with opinion and aspect pairs.}
\label{fig:musicology:annotatedreview}
\end{figure}

For a review aspect $A_{j}$ we determine if there are any sentiment words in the sentence containing $A_{j}$. If not, $A_{j}$ is marked neutral; otherwise, we identify the sentiment word $w_{min}$ with the minimum word-distance to $A_j$. Next, we determine the part-of-speech tags for $w_{min}$, $A_i$ and any words that occur between $w_{min}$ and $A_i$. 
We assign a sentiment score between -1 and 1 to $A_j$ based on the sentiment of $w_{min}$, subject to whether the corresponding sentence contains any negation terms within $4$ words of $w_{min}$. If there are no negation terms, then the sentiment assigned to $A_j$ is that of the sentiment word in the sentiment lexicon; otherwise, this sentiment is reversed. Our sentiment lexicon is derived from SentiWordNet \citep{esuli2006sentiwordnet} and is not specifically tuned for music reviews. 
An overview of the process is shown in Figure~\ref{fig:musicology:OMF}. The end result of sentiment analysis is that we determine a sentiment score $S_{ij}$ for each aspect $A_j$ in review $R_i$. A sample annotated review is shown in Figure~\ref{fig:musicology:annotatedreview}.
Finally, the sentiment score of a review $R_i$ is calculated as the average of the sentiment score $S_{ij}$ of every aspect $A_j$ in $R_i$.

% Figure 9

\subsection{Diachronic study of music criticism}
\label{sec:musicology:evolution}

\begin{figure}
    \centering
    \includegraphics[width=0.60\linewidth]{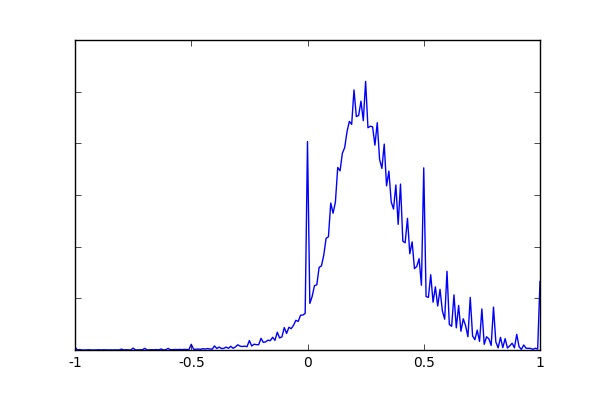}
	\caption{Distribution of sentiment scores}
    \label{fig:musicology:sent-dist}
\end{figure}

We applied the proposed aspect-based sentiment analysis framework to the corpus of album customer reviews gathered from Amazon (see Section~\ref{sec:musicology:mard}), obtaining specific sentiment scores for different aspects present in the text, e.g., album cover, guitar, voice or lyrics. 
In Figure~\ref{fig:musicology:sent-dist} we observe that the sentiment scores follow a Gaussian distribution, with a median of 0.21, and remarkable picks at 0 and 0.5.

% Figure 10

In addition to the sentiment computed, this corpus includes music metadata such as genre, review rating, review publication date and album release date. We benefit from this substantial amount of information at our disposal for performing a diachronic analysis of music criticism. Specifically, we combine the metadata retrieved for each review with their associated sentiment information, and generate visualizations to help us investigate any potential trends in diachronic music appreciation and criticism. Based on this evidence, and since music evokes emotions through mechanisms that are not unique to music \citep{Juslin2008}, we may go as far as using musical information as means for a better understanding of global affairs. Previous studies argue that national confidence may be expressed in any form of art, including music \citep{Moisi2010}, and in fact, there is strong evidence suggesting that our emotional reactions to music have important and far-reaching implications for our beliefs, goals and actions, as members of social and cultural groups \citep{Alcorta2008}. 

To investigate this matter, we carried out a study of the evolution of music criticism from two different temporal standpoints. Specifically, we consider when the review was written and, in addition, when the album was first published. We define the sentiment score of a review as the average score of all aspects in the review. Since we have sentiment information available for each review, we first computed an average sentiment score for each year of review publication (between 2000 and 2014). In this way, we may detect any significant fluctuation in the evolution of affective language during the 21st century. Then, we also calculated an average sentiment score by year of album publication. The affective information is complemented with the averages of the Amazon rating scores.

In what follows, we show visualizations for sentiment scores and correlation with ratings given by Amazon users, according to these two different temporal dimensions. Although arriving to musicological conclusions is out of the scope of this paper, we provide \textit{food for thought} and present the readers with hypotheses that may explain some of the facts revealed by these data-driven trends.

\subsubsection{Evolution by review publication year}
\label{sec:musicology:evolution-review}

% Figure 11
\begin{figure}[ht!]
    \centering    
    \begin{subfigure}{.49\textwidth}
        \centering
        \includegraphics[width=\linewidth]{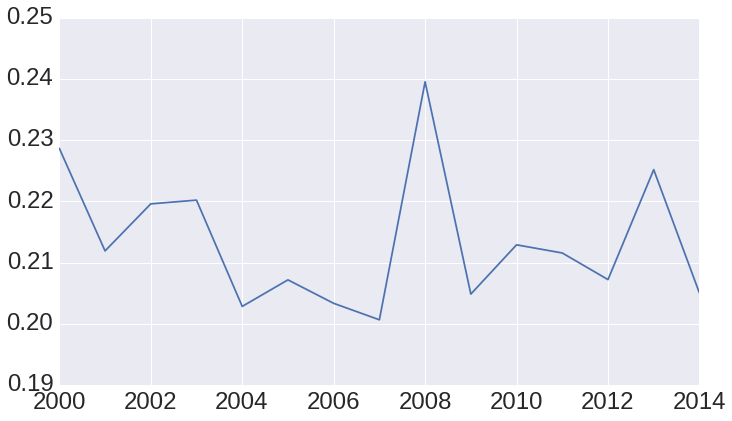}
        \caption{Sentiment}
        \label{fig:musicology:avgSentReview}
    \end{subfigure}
    \begin{subfigure}{.49\textwidth}
        \centering
        \includegraphics[width=\linewidth]{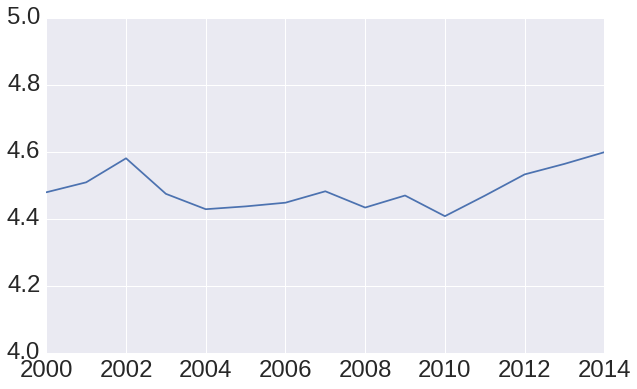}
        \caption{Rating}
        \label{fig:musicology:avgRatingReview}
    \end{subfigure}
    \begin{subfigure}{.49\textwidth}
        \centering
        \includegraphics[width=\linewidth]{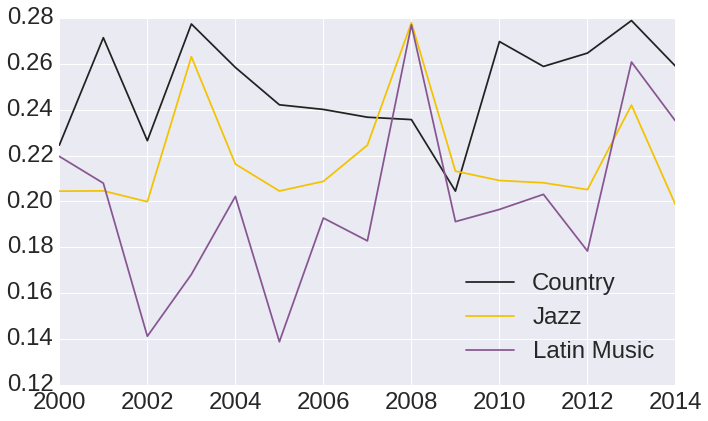}
        \caption{Sentiment by genre}
        \label{fig:musicology:avgSentReviewGenres}
    \end{subfigure}
    \begin{subfigure}{.50\textwidth}
        \centering
        \includegraphics[width=\linewidth]{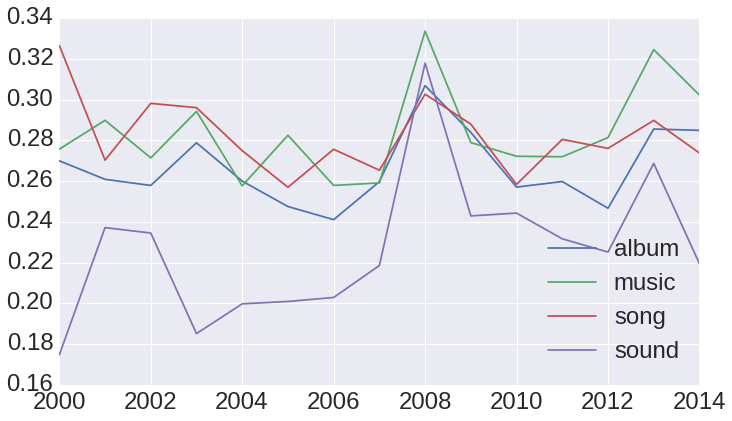}
        \caption{Sentiment by aspect}
        \label{fig:musicology:avgSentReviewAspects}
    \end{subfigure}
    \begin{subfigure}{.50\textwidth}
        \centering
        \includegraphics[width=\columnwidth]{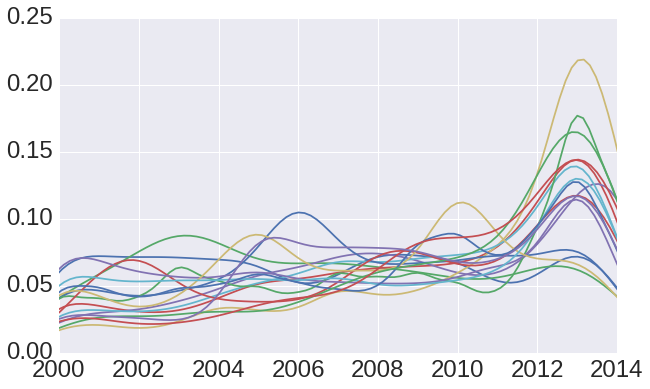}
        \caption{Kernel density est.}
        \label{fig:musicology:kde}
    \end{subfigure}
    \begin{subfigure}{.48\textwidth}
        \centering
        \includegraphics[width=\linewidth]{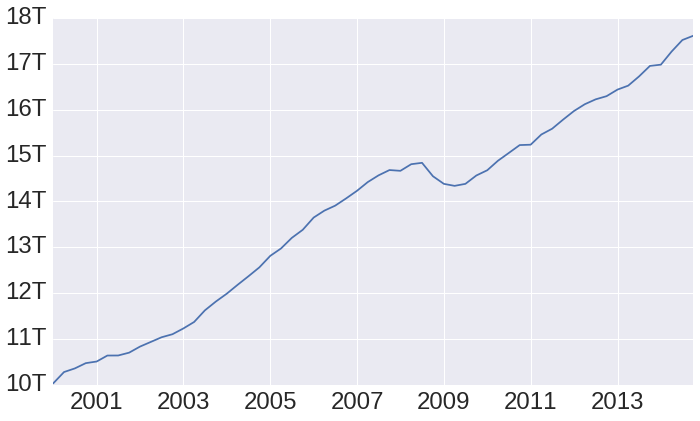}
        \caption{USA GDP trend}
        \label{fig:musicology:gdp}
    \end{subfigure}
    \caption{Sentiment (a, c, and d) and rating (b) averages by review publication year; Kernel density estimation of the distribution of reviews by year (e); GDP trend in USA from 2000 to 2014 (f)}
    \label{fig:musicology:publication-date}
\end{figure}

We applied sentiment and rating average calculations to the whole dataset, grouping album reviews by year of publication of the review. Figure~\ref{fig:musicology:avgSentReview} shows the average of the sentiment scores of all the reviews published in a specific year, whilst Figure~\ref{fig:musicology:avgRatingReview}  shows average review ratings per year. At first sight, we do not observe any correlation between the trends illustrated in the figures. However, the sentiment curve (Figure~\ref{fig:musicology:avgSentReview} ) shows a remarkable peak in 2008, a slightly lower one in 2013, and a low between 2003 and 2007, and also between 2009 and 2012. Figure~\ref{fig:musicology:kde} shows the kernel density estimation of the distribution of reviews by year of the 16 genres. The shapes of these curves suggest that the 2008 peak in the sentiment score is not related to the number of reviews published that year. The peak persists if we construct the graphs with the average sentiment associated with the most repeated aspects in text (Figure~\ref{fig:musicology:avgSentReviewAspects}). 
It is not trivial to give a proper explanation of this variations on the average sentiment. We speculate that these curve fluctuations may suggest some influence of economical or geopolitical circumstances in the language used in the reviews, such as the 2008 election of Barack Obama as president of the US. As stated by the political scientist Dominique Mo\"{i}si in \cite{Moisi2010}:

\begin{displayquote}\small{
In November 2008, at least for a time, hope prevailed over fear. The wall of racial prejudice fell as surely as the wall of oppression had fallen in Berlin twenty years earlier [...] Yet the emotional dimension of this election and the sense of pride it created in many Americans must not be underestimated.}
\end{displayquote}

If we calculate the sentiment evolution curve for the different genres (see Figure~\ref{fig:musicology:avgSentReviewGenres}), we observe that 2008 constitutes an all-time-high for almost all genres. It is remarkable that genres traditionally related to more diverse communities such as Jazz and Latin Music experience such an increase, whilst other genres such as Country do not.

Another factor that might be related to the positiveness in use of language is the economical situation. After several years of continuous economic growth, in 2007 a global economic crisis started\footnote{\url{https://research.stlouisfed.org}}, whose consequences were visible in the society after 2008 (see Figure~\ref{fig:musicology:gdp}). In any case, further study of the different implied variables is necessary to reinforce any of these hypotheses.

\subsubsection{Evolution by album publication year}
\label{sec:musicology:evolution-album}

% Figure 12
\begin{figure}
\centering
   \begin{subfigure}{.49\textwidth}
        \centering
        \includegraphics[width=\textwidth]{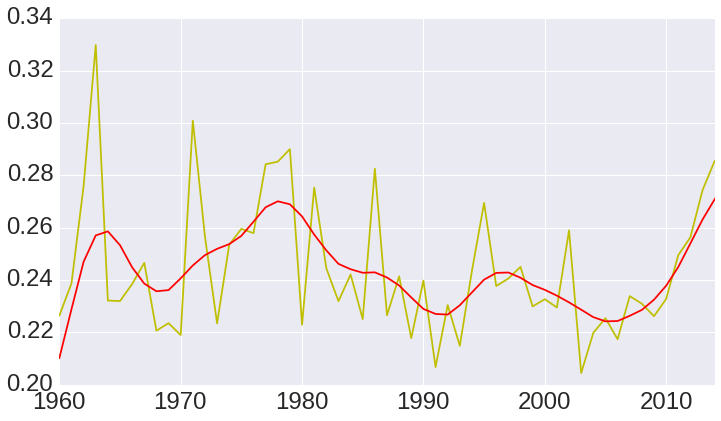}
        \caption{Sentiment}
        \label{fig:musicology:avgSentimentRelease}
    \end{subfigure}
    \begin{subfigure}{.49\textwidth}
        \centering
        \includegraphics[width=\textwidth]{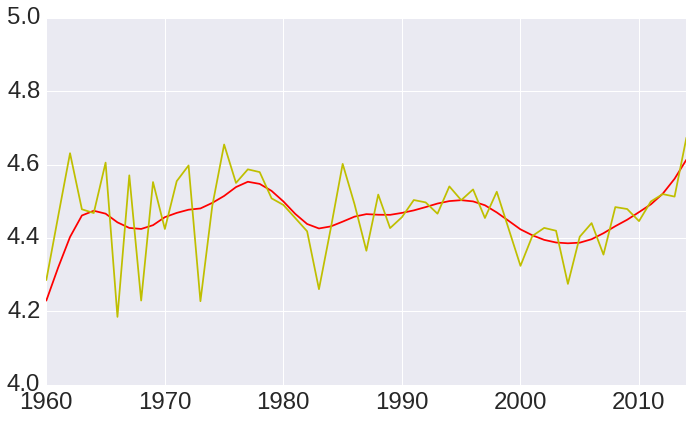}
        \caption{Rating}
        \label{fig:musicology:avgRatingRelease}
    \end{subfigure}  
    \begin{subfigure}{.49\textwidth}
        \centering
        \includegraphics[width=\columnwidth]{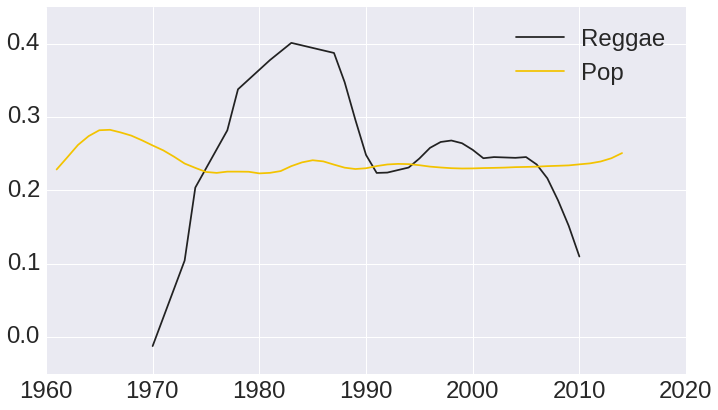}
        \caption{Sentiment by genre}
        \label{fig:musicology:avgSentimentGenresRelease}
    \end{subfigure}
    \caption{Sentiment (a), rating (b), and sentiment by genres (c) averages by album publication year.}
    \label{fig:musicology:release-date}
\end{figure}

In this case, we study the evolution of the polarity of language by grouping reviews according to the album publication date. This date was gathered from MusicBrainz, meaning that this study is conducted on the ~42,1\% of the dataset that was successfully mapped. We compared again the evolution of the average sentiment polarity (Figure~\ref{fig:musicology:avgSentimentRelease}) with the evolution of the average rating (Figure~\ref{fig:musicology:avgRatingRelease}). Contrary to the results observed by review publication year, here we observe a strong correlation between ratings and sentiment polarity. To corroborate that, we computed first a smoothed version of the average graphs, by applying 1-D convolution (see line in red in Figures~\ref{fig:musicology:avgSentimentRelease} and \ref{fig:musicology:avgRatingRelease}). Then we computed Pearson's correlation between smoothed curves, obtaining a correlation $r = 0.75$, and a p-value $p \ll 0.001$. This means that in fact there is a strong correlation between the polarity identified by the sentiment analysis framework in the review texts, and the rating scores provided by the users. This correlation reinforces the conclusions that may be drawn from the sentiment analysis data. 

To further dig into the utility of this polarity measure for studying genre evolution, we also computed the smoothed curve of the average sentiment by genre, and illustrate it with two idiosyncratic genres, namely \textit{Pop} and \textit{Reggae} (see Figure~\ref{fig:musicology:avgSentimentGenresRelease}. We observe in the case of \textit{Reggae} that there is a time period where reviews have a substantial use of a more positive language between the second half of the 70s and the first half of the 80s, an epoch which is often called the golden age of \textit{Reggae} \citep{alleyne2012encyclopedia}. This might be related to the publication of Bob Marley albums, one of the most influential artists in this genre, and the worldwide spread popularity of reggae music. In the case of \textit{Pop}, we observe a more constant sentiment average. However, in the 60s and the beginning of 70s there are higher values, probably consequence by the release of albums by The Beatles and other iconic pop bands. These results show that the use of sentiment analysis on music reviews over certain timelines may be useful to study genre evolution and identify influential events.

Finally, we observe a growing tendency for sentiment and rating average over albums published in the last years of the study. Are we experiencing a new golden age of music or is this due to an increase of the number of reviewers incentivized by 3rd party services~\citep{gibbs_2016}? Again, data-driven analysis has provided us with meaningful insights as well as novel hypotheses that open up vibrating avenues for further studies.

\section{Conclusions}
\label{sec:musicology:conclusions}

We have presented different methodologies to process large corpora of music-related documents from a musicological perspective, enabling data-driven analysis, which can be further used for complementing expert knowledge. These methodologies have been evaluated on three different use cases: Flamenco music (with the creation, population, and analysis of a flamenco knowledge base); the Renaissance period (by processing and studying a corpus of composers biographies); and the evolution of music criticism (with a cross-genre diachronic study using as target data online music reviews).

First, different challenges and techniques for gathering and combining large text corpora are presented and applied to compile the different datasets used throughout the paper. Then, a methodology for the analysis and visualization of word frequencies has been presented, showing some of the most important aspects of the different music schools in Renaissance Music. Next, an information extraction pipeline is presented where entities identified in text are connected to a knowledge base throughout an ad-hoc entity linking system. Then, a method for extracting biographical information from artist biographies is described, and applied to enrich a flamenco knowledge base and to study the Renaissance period. The analysis of extracted biographical data revealed migratory tendencies of composers and the oscillation of the gravitational center of the music activity among different European cities within the Renaissance period. 
A methodology to build knowledge graphs is described next and evaluated in the task of computing artist relevance rankings. Experimental results show high correlation between the automatically obtained ranking of artists and the opinion of flamenco experts. The application on the Renaissance corpus revealed also that the use of only inner connections helps the approach to obtain results that are more musicologically meaningful.
Finally, an aspect-based sentiment analysis method is described. This methodology is then used to perform a diachronic study of the sentiment polarity expressed in customer reviews from two different standpoints. First, an analysis by year of review publication suggests that geopolitical events or macro-economical circumstances may influence the way people speak about music. Second, an analysis by year of album publication shows how sentiment analysis can be useful to study the evolution of music. 

The main contribution of this work is a demonstration of the usefulness for musicological research of applying systematic linguistic processing techniques on text collections \textit{about music}. Although further work is necessary to elaborate on the hypotheses or claims that may be derived from purely data-driven analyses, these methodologies have shown their suitability in the quest of knowledge discovery from large amounts of documents, which may be highly useful for musicologists and humanities researchers in general. In fact, one of the strongest claims used in the Information Age is that \textit{Big Data} can be used to reveal hidden patterns and meaningful variables hidden among unstructured information, and indeed in this work we provide a myriad of conclusions drawn from intelligent text processing, in the hopes that these may constitute the cornerstone of further musicological studies. 
Moreover, we envision that the combination of knowledge extracted from text with knowledge extracted from other data modalities (e.g., audio signals or music scores) would be a further step in the construction of real-world music understanding systems.

\section{Acknowledgments}

This work was partially funded by the Spanish Ministry of Economy and Competitiveness under the Maria de Maeztu Units of Excellence Programme (MDM-2015-0502) and by the COFLA2 research project (Proyectos de Excelencia de la Junta de Andaluc\'{i}a, FEDER P12-TIC-1362).

\bibliographystyle{apacite}
\bibliography{sample.bib}

\end{document}